\documentclass[11pt,letterpaper]{article}
\usepackage{naaclhlt2016}
\usepackage{times}
\usepackage{latexsym}
\usepackage{enumitem}

\makeatletter
\newcommand{\@BIBLABEL}{\@emptybiblabel}
\newcommand{\@emptybiblabel}[1]{}
\makeatother
\usepackage[draft,hidelinks]{hyperref}

\usepackage{color}
\usepackage{amsmath,amsfonts,amssymb,bm}
\usepackage{hyperref}
\usepackage{graphicx,subfigure}
\usepackage[]{algorithm2e}

\renewcommand{\vec}[1]{\boldsymbol{#1}}

\newcommand{\mat}[1]{\mathbf{#1}}

\newcommand{\tuple}[1]{\langle #1 \rangle}

\newcommand{\vz}[0]{\vec{z}}
\newcommand{\vy}[0]{\vec{y}}

\newcommand{\vth}[0]{\vec{\theta}}

\usepackage{adjustbox}
\usepackage{array}
\usepackage{booktabs}

\newcolumntype{R}[2]{%
    >{\adjustbox{angle=#1,lap=\width-(#2)}\bgroup}%
    l%
    <{\egroup}%
}

\newcommand{\rela}[1]{{\sc #1}} 
\newcommand{\word}[1]{``\emph{#1}''} 

\newcommand{\y}[1]{{y}_{#1}} 
\renewcommand{\vy}{\vec{y}}
\newcommand{\h}[1]{\vec{h}_{#1}} 
\newcommand{\z}[1]{{z}_{#1}} 
\newcommand{\cont}[1]{\vec{c}_{#1}} 
\newcommand{\bias}[1]{\vec{b}_{#1}}
\newcommand{\softmax}[1]{\text{softmax}\left(#1\right)}
\newcommand{\func}[1]{\vec{f}_{#1}} 
\newcommand{\outfunc}[1]{g_{#1}} 

\newcommand{\modelone}{{\sc DrLM}} 
\newcommand{\wmat}[1]{\mat{W}_{#1}}
\newcommand{\vmat}[1]{\mat{V}^{(#1)}}
\newcommand{\emat}[1]{\mat{X}_{#1}} 
\newcommand{\mmat}[1]{\mat{M}^{(#1)}}
\newcommand{\jacob}[1]{\textcolor{blue}{(#1 -J)}}
\newcommand{\reza}[1]{\textcolor{green}{(#1 -R)}}
\newcommand{\new}[1]{{#1}}

\naaclfinalcopy 
\def\naaclpaperid{480} 

\title{A Latent Variable Recurrent Neural Network\\for Discourse Relation Language Models}

\author{Yangfeng Ji\\
  Georgia Institute of Technology\\
  Atlanta, GA 30308, USA\\
  {\tt jiyfeng@gatech.edu}
  \And
  Gholamreza Haffari\\
  Monash University\\
  Clayton, VIC, Australia\\
  {\tt gholamreza.haffari}\\
  {\tt @monash.edu}
  \And
  Jacob Eisenstein\\
  Georgia Institute of Technology\\
  Atlanta, GA 30308, USA\\
  {\tt jacobe@gatech.edu}
}

\date{}
\begin{document}

\maketitle

\begin{abstract}
This paper presents a novel latent variable recurrent neural network architecture for jointly modeling sequences of words and (possibly latent) discourse relations between adjacent sentences. A recurrent neural network generates individual words, thus reaping the benefits of discriminatively-trained vector representations. The discourse relations are represented with a latent variable, which can be predicted or marginalized, depending on the task. The resulting model 
can therefore employ a training objective that includes not only discourse relation classification, but also word prediction. As a result, it outperforms state-of-the-art alternatives for two tasks: implicit discourse relation classification in the Penn Discourse Treebank, and dialog act classification in the Switchboard corpus. Furthermore, by marginalizing over latent discourse relations at test time, we obtain a discourse informed language model, which improves over a strong LSTM baseline.
\end{abstract}



\section{Introduction}
\label{sec:intro}
Natural language processing (NLP) has recently
experienced a neural network ``tsunami''~\cite{manning2016computational}. 
A key advantage of these neural architectures is that they employ discriminatively-trained distributed representations, which can capture the meaning of linguistic phenomena ranging from individual words~\cite{turian2010word} to longer-range linguistic contexts at the sentence level~\cite{socher2013recursive} and beyond~\cite{le2014distributed}. Because they are discriminatively trained, these methods can learn representations that yield very accurate predictive models (e.g., Dyer et al, 2015).\nocite{dyer2015transition}

However, in comparison with the probabilistic graphical models that were previously the dominant machine learning approach for NLP, neural architectures lack flexibility. By treating linguistic annotations as random variables, probabilistic graphical models can marginalize over annotations that are unavailable at test or training time, elegantly modeling multiple linguistic phenomena in a joint framework~\cite{finkel2006solving}. But because these graphical models represent uncertainty for every element in the model, adding too many layers of latent variables makes them difficult to train.


In this paper, we present a hybrid architecture that combines a recurrent neural network language model with a latent variable model over shallow discourse structure. In this way, the model learns a discriminatively-trained distributed representation of the local contextual features that drive word choice at the intra-sentence level, using techniques that are now state-of-the-art in language modeling~\cite{mikolov2010recurrent}. However, the model treats shallow discourse structure --- specifically, the relationships between pairs of adjacent sentences --- as a latent variable. As a result, the model can act as both a discourse relation classifier and a language model. Specifically:
\begin{itemize}[itemsep=0pt]
\item If trained to maximize the conditional likelihood of the discourse relations, it outperforms state-of-the-art methods for both implicit discourse relation classification in the Penn Discourse Treebank~\cite{rutherford2015improving} and dialog act classification in Switchboard~\cite{kalchbrenner2013recurrent}. The model learns from both the discourse annotations as well as the language modeling objective, unlike previous recursive neural architectures that learn only from annotated discourse relations~\cite{ji2015one}.
\item If the model is trained to maximize the joint likelihood of the discourse relations and the text, it is possible to marginalize over discourse relations at test time, outperforming language models that do not account for discourse structure.
\end{itemize}

In contrast to recent work on continuous latent variables in recurrent neural networks~\cite{chung2015recurrent}, which require complex variational autoencoders to represent uncertainty over the latent variables, our model is simple to implement and train, requiring only minimal modifications to existing recurrent neural network architectures that are implemented in commonly-used toolkits such as Theano, Torch, and CNN.

We focus on a class of \emph{shallow discourse relations}, which hold between pairs of adjacent sentences (or utterances). These relations describe how the adjacent sentences are related: for example, they may be in \rela{contrast}, or the latter sentence may offer an answer to a question posed by the previous sentence. Shallow relations do not capture the full range of discourse phenomena~\cite{webber2012discourse}, but they account for two well-known problems: implicit discourse relation classification in the Penn Discourse Treebank, which was the 2015 CoNLL shared task~\cite{xue2015conll}; and dialog act classification, which characterizes the structure of interpersonal  communication in the Switchboard corpus~\cite{stolcke2000dialogue}, and is a key component of contemporary dialog systems~\cite{williams2007partially}. 
Our model outperforms state-of-the-art alternatives for implicit discourse relation classification in the Penn Discourse Treebank, and for dialog act classification in the Switchboard corpus.

\section{Background}
\label{sec:background}

Our model scaffolds on recurrent neural network (RNN)  language models~\cite{mikolov2010recurrent}, and recent variants that exploit multiple levels of linguistic detail~\cite{ji2015document,lin2015hierarchical}. 

\paragraph{RNN Language Models} 
Let us denote token $n$ in a sentence $t$ by $\y{t,n} \in \{1 \ldots V\}$, and write $\vy_t = \{ y_{t,n} \}_{n \in \{1\ldots N_{t}\}}$ to indicate the sequence of words in sentence $t$. In an RNN language model, the probability of the sentence is decomposed as,
\begin{small}
  \begin{align}
    p(\vy_t) = & \prod^{N_t}_n p(\y{t,n} \mid \vy_{t,<n}),
  \end{align}
\end{small}
where the probability of each word $\y{t,n}$ is conditioned on the entire preceding sequence of words $\vy_{t,<n}$ through the summary vector $\h{t,n-1}$. This vector is computed recurrently from $\h{t,n-2}$ and from the \emph{embedding} of the current word, 
$\emat{y_{t,n-1}}$, where $\emat{}\in\mathbb{R}^{K\times V}$ and  $K$ is the dimensionality of the word embeddings. The language model can then be summarized as,
\begin{small}
  \begin{align}
    \h{t,n} = & \func{}(\emat{\y{t,n}}, \h{t,n-1})\\
    p(\y{t,n}\mid\vy_{t,<n}) = & \softmax{\wmat{o}\h{t,n-1}+\bias{o}},
  \end{align}
\end{small}
where the matrix $\wmat{o} \in \mathbb{R}^{V \times K}$ defines the \emph{output embeddings}, and $\bias{o} \in \mathbb{R}^V$ is an offset. 
The function $\func{}(\cdot)$ is a deterministic non-linear transition function. 
It typically takes an element-wise non-linear transformation (e.g., $\tanh$) of a vector resulting from the sum of the word embedding and a linear transformation of the previous hidden state. 

The model as described thus far is identical to the recurrent neural network language model (RNNLM) of \newcite{mikolov2010recurrent}. In this paper, we replace the above simple hidden state units with the more complex Long Short-Term Memory units~\cite{hochreiter1997long}, which have consistently been shown to yield much stronger performance in language modeling~\cite{pham2014dropout}. For simplicity, we still use the term RNNLM in referring to this model.


\begin{figure*}
  \centering
  \includegraphics[width=.95\textwidth]{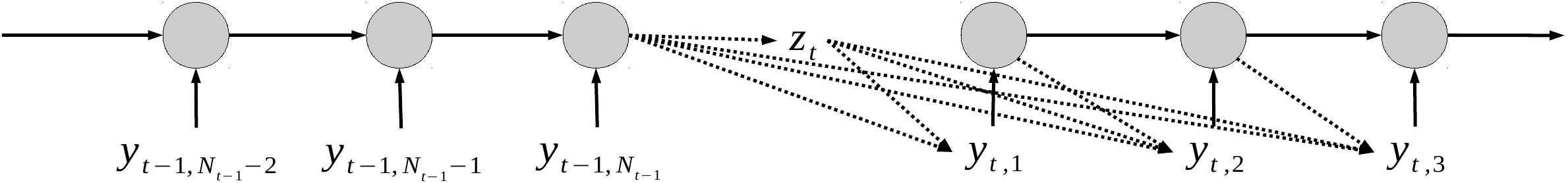}
  \caption{\new{A fragment of our model with latent variable $\z{t}$, which only illustrates discourse information flow from sentence $(t-1)$ to $t$. The information from sentence $(t-1)$ affects the distribution of $\z{t}$ and then the words prediction within sentence $t$.}}
  \label{fig:lvrnn}
\end{figure*}


\begin{figure*}
\begin{equation}
  \label{eq:prob-y-t}
    p(\y{t,n+1}\mid\z{t},~\vy_{t,<n},~\vy_{t-1})=\outfunc{}\Big(\underbrace{\wmat{o}^{(\z{t})}\h{t,n}}_{\begin{array}{c}\text{\scriptsize relation-specific}\\ \text{\scriptsize intra-sentential context}\end{array}} 
    + \underbrace{\wmat{c}^{(\z{t})}\cont{t-1}}_{\begin{array}{c}\text{\scriptsize relation-specific}\\ \text{\scriptsize inter-sentential context}\end{array}}
    + \underbrace{\bias{o}^{(\z{t})}}_{\begin{array}{c}\text{\scriptsize relation-specific}\\ \text{\scriptsize bias}\end{array}}
    \Big)
\end{equation}
\caption{Per-token generative probabilities in the discourse relation language model}
\label{fig:py-drlm}
\end{figure*}

\paragraph{Document Context Language Model}
One drawback of the RNNLM is that it cannot propagate long-range information between the sentences. Even if we remove sentence boundaries,
long-range information will be attenuated by repeated application of the non-linear transition function. 
\newcite{ji2015document} propose the Document Context Language Model (DCLM) to address this issue. The core idea is to represent context with \emph{two} vectors: $\h{t,n}$, representing \emph{intra}-sentence word-level context, and $\cont{t}$, representing \emph{inter}-sentence context. These two vectors are then linearly combined in the generation function for word $\y{t,n}$,
\begin{align}
\notag
 & p(\y{t,n}\mid\vy_{t,<n},\vy_{<t}) \\
& = \softmax{\wmat{o}\h{t,n-1}+\wmat{c}\cont{t-1} + \bias{o}},
\label{eq:dclm}
\end{align}
where $\cont{t-1}$ is set to the last hidden state of the previous sentence. 
\newcite{ji2015document} show that this model can improve language model perplexity.

\section{Discourse Relation Language Models}
\label{sec:model}
We now present a probabilistic neural model over sequences of words and shallow discourse relations. Discourse relations $z_t$ are treated as latent variables, which are linked with a recurrent neural network over words in a \emph{latent variable recurrent neural network}~\cite{chung2015recurrent}.

\subsection{The Model}
Our model  (see Figure \ref{fig:lvrnn}) is formulated as a two-step generative story.
In the first step, context information from the sentence $(t-1)$ 
is used to generate the discourse relation between sentences $(t-1)$ and $t$, 
\begin{equation}
  \label{eq:drlm-pz}
  {\small   p(\z{t}\mid\vy_{t-1})=\softmax{\bm{U}\cont{t-1} + \bias{}},}
\end{equation}
where $\z{t}$ is a random variable capturing the discourse relation between the two sentences, and $\cont{t-1}$ is a vector summary of the contextual information from sentence $(t-1)$, just as in the DCLM (\autoref{eq:dclm}).
\new{The model maintains a default context vector $\cont{0}$ for the first sentences of documents, and treats it as a parameter learned with other model parameters during training.}

In the second step, the sentence $\vy_t$ is generated, conditioning on the preceding sentence $\vy_{t-1}$ and the discourse relation $\z{t}$:
\begin{align}
  \label{eq:drlm-py}
    p(\vy_t \mid \z{t}, \vy_{t-1}) = & \prod_n^{N_t} p(\y{t,n} \mid \vy_{t,<n}, \vy_{t-1}, z_t),
\end{align}
The generative probability for the sentence $\vy_t$ decomposes across tokens as usual (\autoref{eq:drlm-py}). The per-token probabilities are shown in \autoref{eq:prob-y-t}, in \autoref{fig:py-drlm}. Discourse relations are incorporated by parameterizing the output matrices $\wmat{o}^{(\z{t})}$ and $\wmat{c}^{(\z{t})}$; depending on the discourse relation that holds between $(t-1)$ and $t$, these matrices will favor different parts of the embedding space. The bias term $\bias{o}^{(\z{t})}$ is also parametrized by the discourse relation, so that each relation can favor specific words.

Overall, the joint probability of the text and discourse relations is,
\begin{small}
  \begin{align}
    \label{eq:drlm-joint-likelihood}
    p(\vy_{1:T}, \vz_{1:T}) = & \prod_t^T p(\z{t} \mid \vy_{t-1})\times p(\vy_t \mid \z{t}, \vy_{t-1}).
  \end{align}
\end{small}
If the discourse relations $z_t$ are not observed, then our model is a form of latent variable recurrent neural network (LVRNN). Connections to recent work on LVRNNs are discussed in \autoref{sec:related}; the key difference is that the latent variables here correspond to linguistically meaningful elements, which we may wish to predict or marginalize, depending on the situation.

\paragraph{Parameter Tying}
As proposed, the Discourse Relation Language Model has a large number of parameters. Let $K$, $H$ and $V$ be the input dimension, hidden dimension and the size of vocabulary in language modeling. The size of each prediction matrix $\wmat{o}^{(\z{})}$ and $\wmat{c}^{(\z{})}$ is $V\times H$; there are two such matrices for each possible discourse relation. We reduce the number of parameters by factoring each of these matrices into two components:
\begin{align}
  \label{eq:factorization}
  \wmat{o}^{(\z{})} = \wmat{o}\cdot\vmat{\z{}}, \quad  \wmat{c}^{(\z{})} = \wmat{c}\cdot\mmat{\z{}},
\end{align}
where $\vmat{\z{}}$ and $\mmat{\z{}}$ are relation-specific components for intra-sentential and inter-sentential contexts; the size of these matrices is $H\times H$, with $H \ll V$. The larger $V\times H$ matrices $\wmat{o}$ and $\wmat{c}$ are shared across all relations.


\subsection{Inference}
\label{sec:inference}
There are two possible inference scenarios: inference over discourse relations, conditioning on words; and inference over words, marginalizing over discourse relations.

\paragraph{Inference over Discourse Relations}
The probability of discourse relations given the sentences $ p(\vz_{1:T} \mid \vy_{1:T})$ is decomposed into the product of probabilities of individual discourse relations conditioned on the adjacent sentences $\prod_t p(\z{t} \mid \vy_t, \vy_{t-1})$. These probabilities are computed by Bayes' rule:

\begin{small}
  \begin{equation}
    \label{eq:post-z}
    p(\z{t} \mid \vy_t, \vy_{t-1}) = \frac
    {p(\vy_t \mid \z{t}, \vy_{t-1}) \times p(\z{t} \mid \vy_{t-1})}
    {\sum_{z'} p(\vy_t \mid z', \vy_{t-1}) \times p(z' \mid \vy_{t-1})}.
  \end{equation}
\end{small}
The terms in each product are given in Equations~\ref{eq:drlm-pz} and~\ref{eq:drlm-py}. Normalizing involves only a sum over a small finite number of discourse relations. Note that inference is easy in our case because all words are observed and there is no probabilistic coupling of the discourse relations.

\paragraph{Inference over Words} 
In discourse-informed language modeling, we marginalize over discourse relations to compute the probability of a sequence of sentence $\vy_{1:T}$, which can be written as,
\begin{equation}
  \label{eq:drlm-marginalized-likelihood}
{\small p(\vy_{1:T}) 
  = \prod_t^T \sum_{\z{t}} p(\z{t}\mid\vy_{t-1})\times p(\vy_t\mid\z{t}, \vy_{t-1}),}
 \end{equation}
because the word sequences are observed, decoupling each $\z{t}$ from its neighbors $\z{t+1}$ and $\z{t-1}$. This decoupling ensures that we can compute the overall marginal likelihood as a product over local marginals.

\subsection{Learning}
\label{sec:learn}
The model can be trained in two ways: to maximize the joint probability $p(\vy_{1:T}, \vz_{1:T})$, or to maximize the conditional probability $p(\vz_{1:T} \mid \vy_{1:T})$. The joint training objective is more suitable for language modeling scenarios, and the conditional objective is better for discourse relation prediction. We now describe each objective in detail.

\paragraph{Joint likelihood objective}
The joint likelihood objective function is directly adopted from the joint probability defined in \autoref{eq:drlm-joint-likelihood}. The objective function for a single document with $T$ sentences or utterances is,
\begin{small}
  \begin{align}
    \notag
    \ell(\bm{\theta}) = & \sum_t^T \log p(z_t \mid \vy_{t-1}) \\
                        & + \sum_n^{N_t} \log p(\y{t,n} \mid \vy_{t,<n}, \vy_{t-1}, \z{t}),
    \label{eq:obj-joint}
  \end{align}
\end{small}
where $\vth$ represents the collection of all model parameters, including the parameters in the LSTM units and the word embeddings.

Maximizing the objective function $\ell(\bm{\theta})$ will jointly optimize the model on both language language and discourse relation prediction. As such, it can be viewed as a form of multi-task learning~\cite{caruana1997multitask}, where we learn a shared representation that works well for discourse relation prediction and for language modeling. However, in practice, the large vocabulary size and number of tokens means that the language modeling part of the objective function tends to dominate.

\paragraph{Conditional objective}
This training objective is specific to the  
discourse relation prediction task, and  based on \autoref{eq:post-z} can be written as: 
%
\begin{equation}
  \label{eq:obj-rela}
  {\small 
    \begin{split}
      \ell_r(\vth) = & \sum_t^T \log p(z_t \mid \vy_{t-1}) + \log p(\vy_t \mid z_t, \vy_{t-1})\\
      & - \log \sum_{z'} p(z' \mid \vy_{t-1}) \times p(\vy_t \mid z', \vy_{t-1})
  \end{split}}
\end{equation}

The first line in \autoref{eq:obj-rela} is the same as $\ell(\vth)$,
but the second line reflects the normalization over all possible values of $\z{t}$. \new{This forces the objective function to attend specifically to the problem of maximizing the conditional likelihood of the discourse relations and treat language modeling as an auxiliary task \cite{collobert2011natural}}.

\subsection{Modeling limitations} 
\label{sec:model-limitations}
The discourse relation language model is carefully designed to decouple the discourse relations from each other, after conditioning on the words. It is clear that text documents and spoken dialogues have sequential discourse structures, and it seems likely that modeling this structure could improve  performance. In a traditional hidden Markov model (HMM) generative approach~\cite{stolcke2000dialogue}, modeling sequential dependencies is not difficult, because training reduces to relative frequency estimation. However, in the hybrid probabilistic-neural architecture proposed here, training is already expensive, due to the large number of parameters that must be estimated. Adding probabilistic couplings between adjacent discourse relations $\tuple{z_{t-1}, z_t}$ would require the use of dynamic programming for both training and inference, increasing time complexity by a factor that is quadratic in the number of discourse relations. We did not attempt this in this paper; we do compare against a conventional HMM on the dialogue act prediction task in \autoref{sec:exp}. 

\newcite{ji2015document} propose an alternative form of the document context language model, in which the contextual information $\cont{t}$ impacts the hidden state $\h{t+1}$, rather than going directly to the outputs $\vy_{t+1}$. They obtain slightly better perplexity with this approach, which has fewer trainable parameters. However, this model would couple $z_t$ with \emph{all} subsequent sentences $\vy_{>t}$, making prediction and marginalization of discourse relations considerably more challenging. Sequential Monte Carlo algorithms offer a possible solution~\cite{de2000sequential,gu2015neural}, which may be considered in future work.





\section{Data and Implementation}
\label{sec:data}

We evaluate our model on two benchmark datasets: (1) the Penn Discourse Treebank \cite[PDTB]{prasad2008penn}, which is annotated on a corpus of Wall Street Journal acticles; (2) the Switchboard dialogue act corpus \cite[SWDA]{stolcke2000dialogue}, which is annotated on a collections of phone conversations. Both corpora contain annotations of discourse relations and dialogue relations that hold between adjacent spans of text. 

\newcommand{\argone}[0]{\texttt{Arg1}}
\newcommand{\argtwo}[0]{\texttt{Arg2}}
\paragraph{The Penn Discourse Treebank (PDTB)} provides a low-level discourse annotation on written texts. In the PDTB, each discourse relation is annotated between two argument spans, \argone\ and \argtwo. There are two types of relations: explicit and implicit. 
Explicit relations are signalled by discourse markers (e.g., \word{however}, \word{moreover}), and the span of \argone\ is almost totally unconstrained: it can range from a single clause to an entire paragraph, and need not be adjacent to either \argtwo\ nor the discourse marker. However, automatically classifying these relations is considered to be relatively easy, due to the constraints from the discourse marker itself~\cite{pitler2008easily}.
\new{In addition, explicit relations are difficult to incorporate into language models which must generate each word exactly once.}
\new{On the contrary}, \emph{implicit} discourse relations are annotated only between adjacent sentences, based on a semantic understanding of the discourse arguments. Automatically classifying these discourse relations is a challenging task~\cite{lin2009recognizing,pitler2009automatic,rutherford2015improving,ji2015one}. We therefore focus on implicit discourse relations, leaving to the future work the question of how to apply our modeling framework to explicit discourse relations. During training, we collapse all relation types other than implicit (explicit, \textsc{EntRel}, and \textsc{NoRel}) into a single dummy relation type, which holds between all adjacent sentence pairs that do not share an implicit relation.



As in the prior work on first-level discourse relation identification (e.g., Park and Cardie, 2012\nocite{park2012implicit}), we use sections 2-20 of the PDTB as the training set, sections 0-1 as the development set for parameter tuning, and sections 21-22 for testing. 
For preprocessing, we lower-cased all tokens, and substituted all numbers with a special token \word{NUM}. To build the vocabulary, we kept the 10,000 most frequent words from the training set, and replaced low-frequency words with a special token \word{UNK}. 
In prior work that focuses on detecting individual relations, balanced training sets are constructed so that there are an equal number of instances with and without each relation type~\cite{park2012implicit,biran2013aggregated,rutherford2014discovering}. In this paper, we target the more challenging multi-way classification problem, so this strategy is not applicable; in any case, since our method deals with entire documents, it is not possible to balance the training set in this way.

\paragraph{The Switchboard Dialog Act Corpus (SWDA)} is annotated on the Switchboard Corpus of human-human conversational telephone speech~\cite{godfrey1992switchboard}. The annotations label each utterance with one of 42 possible speech acts, such as \rela{agree}, \rela{hedge}, and \rela{wh-question}. Because these speech acts form the structure of the dialogue, most of them pertain to both the preceding and succeeding utterances (e.g., \rela{agree}). The SWDA corpus includes 1155 five-minute conversations. 
We adopted the standard split from \newcite{stolcke2000dialogue}, using 1,115 conversations for training and nineteen conversations for test. For parameter tuning, we randomly select nineteen conversations from the training set as the development set. After parameter tuning, we train the model on the full training set with the selected configuration. We use the same preprocessing techniques here as in the PDTB.


\subsection{Implementation}
\label{subsec:imp}
We use a single-layer LSTM to build the recurrent architecture of our models, which we implement in the {\sc cnn} package.\footnote{\url{https://github.com/clab/cnn}}
\new{Our implementation is available on \url{https://github.com/jiyfeng/drlm}.}
Some additional details follow.

\paragraph{Initialization} Following prior work on RNN initialization~\cite{bengio2012practical}, all parameters except the relation prediction parameters $\mat{U}$ and $\bias{}$ are initialized with random values drawn from the range $[-\sqrt{6/(d_1+d_2)},\sqrt{6/(d_1+d_2)}]$, where $d_1$ and $d_2$ are the input and output dimensions of the parameter matrix respectively. 
The matrix $\mat{U}$ is initialized with random numbers from $[-10^{-5},10^{-5}]$ and $\bias{}$ is initialized to $\bm{0}$.

\paragraph{Learning} Online learning was performed using AdaGrad~\cite{duchi2011adaptive} with initial learning rate $\lambda=0.1$. To avoid the exploding gradient problem, we used  norm clipping trick with a threshold of $\tau=5.0$~\cite{pascanu2012difficulty}. In addition, we used value dropout~\cite{srivastava2014dropout} with rate $0.5$, on the input $\emat{}$, context vector $\cont{}$ and hidden state $\h{}$, similar to the architecture proposed by \newcite{pham2014dropout}.
\new{The training procedure is monitored by the performance on the development set. In our experiments, 4 to 5 epochs were enough.}

\paragraph{Hyper-parameters} Our model includes two tunable hyper-parameters: the dimension of word representation $K$, the hidden dimension of LSTM unit $H$. We consider the values $\{32, 48, 64, 96,128\}$ for both $K$ and $H$. For each corpus in experiments, the best combination of $K$ and $H$ is selected via grid search on the development set.



\section{Experiments}
\label{sec:exp}
Our main evaluation is discourse relation prediction, using the PDTB and SWDA corpora. We also evaluate on language modeling, to determine whether incorporating discourse annotations at training time and then marginalizing them at test time can improve performance.

\subsection{Implicit discourse relation prediction on the PDTB}
\label{subsec:implicit}
We first evaluate our model with implicit discourse relation prediction on the PDTB dataset. Most of the prior work on first-level discourse relation prediction focuses on the ``one-versus-all'' binary classification setting, but we attack the more general four-way classification problem, as performed by \newcite{rutherford2015improving}. We compare against the following methods: 
\begin{description}[style=unboxed,leftmargin=0.4cm,itemsep=0cm]
\item[\newcite{rutherford2015improving}] build a set of feature-rich classifiers on the PDTB, and then augment these classifiers with additional automatically-labeled training instances. We compare against their published results, which are state-of-the-art.
\item[\newcite{ji2015one}] employ a recursive neural network architecture. Their experimental setting is different, so we re-run their system using the same setting as described in \autoref{sec:data}.
\end{description}



\begin{table}
  \centering
  {\small
  \begin{tabular}{lll}
    \toprule
    Model & Accuracy & Macro $F_1$\\
    \midrule
    {\em Baseline}\\
    1. Most common class & 54.7 &--- \\[0.5em]
    {\em Prior work}\\
    2. \cite{rutherford2015improving} & 55.0 & 38.4\\
    3. \cite{rutherford2015improving} & 57.1 & 40.5\\ 
    ~~~~~with extra training data & & \\
    4. \cite{ji2015one} & 56.4 & 40.0\\[0.3em]
    {\em Our work - \modelone}\\
    5. Joint training & 57.1 & 40.5\\
    6. Conditional training & {\bf 59.5}$^\ast$ & {\bf 42.3}\\[0.3em]
    \bottomrule
    \multicolumn{3}{l}{\scriptsize $^\ast$ significantly better than lines 2 and 4 with $p < 0.05$}\\
  \end{tabular}}
  \caption{Multiclass relation identification on the first-level PDTB relations.}
  \label{tab:rela-levelone}
\end{table}

\paragraph{Results}
As shown in \autoref{tab:rela-levelone}, the conditionally-trained discourse relation language models (\modelone) outperforms all alternatives, on both metrics. \new{While the jointly-trained \modelone\ is at the same level as the previous state-of-the-art, conditional training on the same model provides a significant additional advantage, indicated by a binomial test.}

\subsection{Dialogue Act tagging}
\label{subsec:dialogue}
Dialogue act tagging has been widely studied in both NLP and speech communities. We follow the setup used by~\newcite{stolcke2000dialogue} to conduct experiments, and adopt the following systems for comparison:
\begin{description}[style=unboxed,leftmargin=0.4cm,itemsep=0cm]
\item[\newcite{stolcke2000dialogue}] employ a hidden Markov model, with each HMM state corresponding to a dialogue act.
\item[\newcite{kalchbrenner2013recurrent}] employ a complex neural architecture, with a convolutional network at each utterance and a recurrent network over the length of the dialog. To our knowledge, this model attains state-of-the-art accuracy on this task, outperforming other prior work such as~\cite{webb2005dialogue,milajevs2014investigating}.
\end{description}

\paragraph{Results}
As shown in \autoref{tab:rela-swda}, \new{the conditionally-trained discourse relation language model (\modelone) outperforms all competitive systems on this task. A binomial test shows the result in line 6 is significantly better than the previous state-of-the-art (line 4).} All comparisons are against published results, and Macro-$F_1$ scores are not available. Accuracy is more reliable on this evaluation, since no single class dominates, unlike the PDTB task.


\begin{table}
  \centering
  {
  \begin{tabular}{p{5.3cm}l}
    \toprule
    1. Model & Accuracy\\
    \midrule
    {\em Baseline}\\
    2. Most common class & 31.5 \\[0.3em]
    {\em Prior work}\\
    3. \cite{stolcke2000dialogue} & 71.0\\
    4. \cite{kalchbrenner2013recurrent} & 73.9\\[0.3em]
    {\em Our work - \modelone}\\
    5. Joint training & 74.0 \\
    6. Conditional training & {\bf 77.0}$^\ast$ \\
    \bottomrule
    \multicolumn{2}{l}{\scriptsize $^\ast$ significantly better than line 4 with $p < 0.01$}
  \end{tabular}}
  \caption{The results of dialogue act tagging.}
  \label{tab:rela-swda}
\end{table}

\subsection{Discourse-aware language modeling}
\label{subsec:language}
As a joint model for discourse and language modeling, \modelone\ can also function as a language model, assigning probabilities to sequences of words while marginalizing over discourse relations. To determine whether discourse-aware language modeling can improve performance, we compare against the following systems: 
\begin{description}[style=unboxed,leftmargin=0.4cm,itemsep=0cm]
\item[RNNLM+LSTM] This is the same basic architecture as the RNNLM proposed by \cite{mikolov2010recurrent}, which was shown to outperform a Kneser-Ney smoothed 5-gram model on modeling Wall Street Journal text. Following \newcite{pham2014dropout}, we replace the Sigmoid nonlinearity with a long short-term memory (LSTM).
\item[DCLM] We compare against the Document Context Language Model (DCLM) of \newcite{ji2015document}. We use the ``context-to-output'' variant, which is identical to the current modeling approach, except that it is not parametrized by discourse relations. This model achieves strong results on language modeling for small and medium-sized corpora, outperforming RNNLM+LSTM.
\end{description}

\paragraph{Results} The perplexities of language modeling on the PDTB and the SWDA are summarized in \autoref{tab:ppl}. The comparison between line 1 and line 2 shows the benefit of considering multi-sentence context information on language modeling. Line 3 shows that adding discourse relation information yields further improvements for both datasets. We emphasize that discourse relations in the test documents are marginalized out, so no annotations are required for the test set; the improvements are due to the disambiguating power of discourse relations in the training set.

Because our training procedure requires discourse annotations, this approach does not scale to the large datasets typically used in language modeling. As a consequence, the results obtained here are somewhat academic, from the perspective of practical language modeling. Nonetheless, the positive results here motivate the investigation of training procedures that are also capable of marginalizing over discourse relations at training time.


\begin{table}
  \centering
  \hspace{-3ex}
  {\small
  \begin{tabular}{llllllll}
    \toprule
    & \multicolumn{3}{l}{PDTB} & &\multicolumn{3}{l}{SWDA} \\
    \cmidrule{2-4} \cmidrule{6-8}
    Model & $K$ & $H$ & \textsc{pplx} & & $K$ & $H$ & \textsc{pplx}\\
    \midrule
    {\em Baseline}\\
    1. RNNLM  & 96 & 128 & 117.8 & & 128 & 96 & 56.0\\
    2. DCLM & 96 & 96 & 112.2 & & 96 & 96 & 45.3\\[0.3em]
    {\em Our work}\\
    3. \modelone & 64 & 96 & 108.3 & & 128 & 64 & 39.6 \\
    \bottomrule
  \end{tabular}}
  \caption{Language model perplexities (\textsc{pplx}), lower is better. The model dimensions $K$ and $H$ that gave best performance on the dev set are also shown.}
  \label{tab:ppl}
\end{table}





\section{Related Work}
\label{sec:related}

This paper draws on previous work in both discourse modeling and language modeling. 

\paragraph{Discourse and dialog modeling} 
Early work on discourse relation classification utilizes rich, hand-crafted feature sets~\cite{joty2012novel,lin2009recognizing,sagae2009analysis}.
Recent representation learning approaches attempt to learn good representations jointly with discourse relation classifiers and discourse parsers~\cite{ji2014representation,li2014recursive}. Of particular relevance are applications of neural architectures to PDTB implicit discourse relation classification~\cite{ji2015one,zhang2015shallow,braud2015comparing}. All of these approaches are essentially classifiers, and take supervision only from the 16,000 annotated discourse relations in the PDTB training set. In contrast, our approach is a probabilistic model over the entire text. 

Probabilistic models are frequently used in dialog act tagging, where hidden Markov models have been a dominant approach~\cite{stolcke2000dialogue}. In this work, the emission distribution is an $n$-gram language model for each dialogue act; we use a conditionally-trained recurrent neural network language model. An alternative neural approach for dialogue act tagging is the combined convolutional-recurrent architecture of \newcite{kalchbrenner2013recurrent}. Our modeling framework is simpler, relying on a latent variable parametrization of a purely recurrent architecture.


\paragraph{Language modeling} There are an increasing number of attempts to incorporate document-level context information into language modeling. For example, \newcite{mikolov2012context} introduce LDA-style topics into RNN based language modeling. \new{\newcite{sordoni2015neural} use a convolutional structure to summarize the context from previous two utterances as context vector for RNN based language modeling. Our models in this paper provide a unified framework to model the context and current sentence.} \newcite{wang2015larger} and \newcite{lin2015hierarchical} construct bag-of-words representations of previous sentences, which are then used to inform the RNN language model that generates the current sentence. The most relevant work is the Document Context Language Model~\cite[DCLM]{ji2015document}; we describe the connection to this model in \autoref{sec:background}. By adding discourse information as a latent variable, we attain better perplexity on held-out data.

\paragraph{Latent variable neural networks} 
Introducing latent variables to a neural network model increases its  representational capacity, which is the main goal of prior efforts in this space~\cite{kingma2013auto,chung2015recurrent}. From this perspective, our model with discourse relations as latent variables shares the same merit. Unlike this prior work, in our approach, the latent variables carry a linguistic interpretation, and are at least partially observed. Also, these prior models employ continuous latent variables, requiring complex inference techniques such as variational autoencoders~\cite{kingma2013auto,burda2016importance,chung2015recurrent}. In contrast, the discrete latent variables in our model are easy to sum and maximize over.



\section{Conclusion}
\label{sec:con}
We have presented a probabilistic neural model over sequences of words and shallow discourse relations between adjacent sequences. This model combines positive aspects of neural network architectures with probabilistic graphical models: it can learn discriminatively-trained vector representations, while maintaining a probabilistic representation of the targeted linguistic element: in this case, shallow discourse relations. This method outperforms state-of-the-art systems in two discourse relation detection tasks, and can also be applied as a language model, marginalizing over discourse relations on the test data. Future work will investigate the possibility of learning from partially-labeled training data, which would have at least two potential advantages. First, it would enable the model to scale up to the large datasets needed for competitive language modeling. Second, by training on more data, the resulting vector representations might support even more accurate discourse relation prediction.

\section*{Acknowledgments}
Thanks to Trevor Cohn, Chris Dyer, Lingpeng Kong, and Quoc V. Le for helpful discussions, and to the anonymous reviewers for their feedback. This work was supported by a Google Faculty Research award to the third author. It was partially performed during the 2015 Jelinek Memorial Summer Workshop on Speech and Language Technologies at the University of Washington, Seattle, and was supported by Johns Hopkins University via NSF Grant No IIS 1005411, DARPA LORELEI Contract No HR0011-15-2-0027, and gifts from Google, Microsoft Research, Amazon and Mitsubishi Electric Research Laboratory. 

\bibliographystyle{naaclhlt2016}
\bibliography{cite-strings,cites,cite-definitions}

\end{document}